\documentclass[journal,twoside,web]{ieeecolor}
\usepackage{tmi}
\usepackage{hyperref}
\usepackage{graphicx}
\usepackage{multirow}
\usepackage{adjustbox}
\usepackage{tabularx}
\usepackage{booktabs}
\usepackage{amsfonts}
\usepackage{pifont}
\usepackage{cite}
\usepackage{amsmath,amssymb,amsfonts}
\usepackage{algorithmic}
\usepackage{graphicx}
\usepackage{textcomp}
\def\BibTeX{{\rm B\kern-.05em{\sc i\kern-.025em b}\kern-.08em
    T\kern-.1667em\lower.7ex\hbox{E}\kern-.125emX}}
\begin{document}
\title{Anatomically Constrained Transformers for Echocardiogram Analysis}
\author{Alexander Thorley, Agis Chartsias, Jordan Strom, Jeremy Slivnick, \\ Dipak Kotecha, Alberto Gomez, Jinming Duan
\thanks{Manuscript received X; accepted X. Date of publication X; date of current version X. This work was supported by Ultromics Ltd.}
\thanks{A. Thorley and J. Duan are with the University of Birmingham, School of Computer Science, UK. (e-mail: ajt973@student.bham.ac.uk)}
\thanks{J. Strom is with the Beth Isreal Deaconess Medical Centre, USA.}
\thanks{J. Slivnick is with the University of Chicago, USA.}
\thanks{D. Kotecha is with the Institute of Cardiovascular Sciences, University of Birmingham, UK.}
\thanks{A. Thorley, A. Chartsias and A. Gomez are with Ultromics Ltd, UK.}}
\maketitle
\begin{abstract}
\bstctlcite{IEEEexample:BSTcontrol}
Video transformers have recently demonstrated strong potential for echocardiogram (echo) analysis, leveraging self-supervised pre-training and flexible adaptation across diverse tasks. However, like other models operating on videos, they are prone to learning spurious correlations from non-diagnostic regions such as image backgrounds. To overcome this limitation, we propose the Video Anatomically Constrained Transformer (ViACT), a novel framework that integrates anatomical priors directly into the transformer architecture. ViACT represents a deforming anatomical structure as a point set and encodes both its spatial geometry and corresponding image patches into transformer tokens. During pre-training, ViACT follows a masked autoencoding strategy that masks and reconstructs only anatomical patches, enforcing that representation learning is focused on the anatomical region. The pre-trained model can then be fine-tuned for tasks localized to this region. In this work we focus on the myocardium, demonstrating the framework on echo analysis tasks such as left ventricular ejection fraction (EF) regression and cardiac amyloidosis (CA) detection. The anatomical constraint focuses transformer attention within the myocardium, yielding interpretable attention maps aligned with regions of known CA pathology. Moreover, ViACT generalizes to myocardium point tracking without requiring task-specific components such as correlation volumes used in specialized tracking networks.
\end{abstract}


\section{Introduction}
\IEEEPARstart{E}{chocardiography} is the cornerstone of modern cardiology as a low cost, accessible imaging modality which provides clinicians with information about cardiac structure and function including EF and global longitudinal strain (GLS)~\cite{lang2015recommendations}. Many works have applied deep learning methods to automate tasks in the echo processing workflow such as view classification~\cite{ostvik2019real}, the calculation of clinical metrics~\cite{ostvik2018automatic,ouyang2020video} and directly detecting cardiovascular disease~\cite{akerman2023automated}. 
There has been a recent shift in the literature towards foundation models such as large vision transformers (ViTs), pre-trained via self-supervised learning (SSL) for use on multitude of different tasks~\cite{amadou2024echoapex, echofm}. In this paradigm, a single transformer model is first pre-trained on unlabeled datasets with a pre-text task such as masked image reconstruction~\cite{he2022masked}. The model is then fine tuned for specific tasks on a smaller sets of labeled data, thus removing the need for task specific model architectures with potential to improve performance and generalizability~\cite{amadou2024echoapex}. This paradigm is promising for echo imaging where large sets of unlabeled echos are more obtainable due to the accessibility of the modality, whilst reducing the need for large labeled datasets.
\\
\indent
Most existing works in echo utilizing transformers and SSL, process entire echo videos as per their computer vision counterparts~\cite{amadou2024echoapex, zhang2024echo, echofm}. This provides no guarantee that the model is using clinically relevant image features as opposed to irrelevant information such as burnt in annotations and imaging artifacts~\cite{bransby2024backmix}. Furthermore, pre-trained encoders are typically tuned for only semantic segmentation~\cite{amadou2024echoapex, echofm}, limiting their application to echocardiography where the measurement of regional deformation in cardiac anatomy is necessary for calculating metrics such as regional longitudinal strain (RLS). To calculate myocardial strain, points along a myocardium contour must be tracked through the cardiac cycle and strain values calculated either between neighboring contour points (RLS) or globally along the entire contour (GLS) as it deforms. The task has traditionally been tackled via speckled tracking~\cite{d2016two}, and more recently with specialist optical flow~\cite{ostvik2021myocardial, evain2022motion} and point tracking architectures~\cite{abulkalam2024echotracker, chernyshov2024automated}.
\\
\indent
Inspired by the wealth of prior works explicitly embedding anatomical knowledge of cardiac geometry into traditional machine and deep learning architectures~\cite{sanchez2018machine,upton2022automated, chiou2021ai, hathaway2022ultrasonic}, we propose an anatomically constrained ViACT model to tackle the aforementioned limitations in existing works. Our model processes sets of points parameterizing cardiac anatomy in echo videos, which in this work we restrict to the myocardium, as most common clinical measurements are derived from myocardial structure and motion. These non-integer myocardium points are used to sample patches from the echo frames which are embedded along with the points themselves into tokens for processing with a transformer. This ensures that the model only uses features localized to the myocardium, removing the possibility for the model to utilize irrelevant image content. The model can be tuned for tasks such as left ventricular EF regression and the classification of CA, which has a number of known abnormalities localized to the myocardium~\cite{dorbala2021asnc}. Furthermore, as the model processes point coordinates, it can also be adapted to track a set of myocardium contour points through an echo clip, capturing regional deformations between points over time, which semantic segmentation alone cannot. In contrast to video pre-training with MAE \cite{feichtenhofer2022masked,tong2022videomae}, our ViACT can be pre-trained to reconstruct masked myocardium patches instead of full video clips leading to improved performance across all tasks and a significant reduction in compute requirements. In summary, the main contributions are as follows:
\begin{itemize}
    \item We propose ViACT, an adaption to the ViT, which processes myocardium points and corresponding image patches sampled at non-integer point locations from all frames in an echo clip. This provides a guarantee that the model is focused on the myocardium, enabling us to visualize attention maps along the deforming anatomy. 
    \item Inspired by video pre-training strategies~\cite{tong2022videomae,feichtenhofer2022masked}, we introduce an anatomical MAE framework to mask and reconstruct video sequences of myocardium patches. This greatly reduces the computational requirements of pre-training compared with a video model of the same size.
    \item We show that the pre-trained ViACT is a general backbone model for echo analysis capable of disease classification, EF regression and myocardium point tracking with superior performance compared with the majority of state-of-the-art methods. To our knowledge, this is the first general-purpose pre-trained model to incorporate point tracking in echocardiography, with related works tackling only segmentation and landmark detection~\cite{amadou2024echoapex, echofm}. Our model removes the need for task-specific components, such as correlation volumes used by point tracker architectures~\cite{abulkalam2024echotracker, karaev2024cotracker, chernyshov2024automated}.
\end{itemize}

This paper extends our preliminary work on a space-time factorized variant of the ViACT~\cite{thorley2025anatomically} to a full video model and pre-training framework. We expand on preliminary experiments tuning ViACTs for disease classification to also include EF regression and point tracking, demonstrating the adaptability of the model for multiple echo analysis tasks.
\section{Related Work}
\subsection{Transformers in Echocardiography}
Early work from~\cite{reynaud2021ultrasound} used a BERT transformer to aggregate frame-wise convolutional neural network (CNN) features from differing length clips to predict ejection fraction (EF). This style of space-time factorization has proved popular in the literature, where each frame in a video is independently embedded with a frame encoder and the resultant embeddings aggregated via a separate temporal transformer. Many works have followed a space-time factorized ViViT architecture~\cite{mokhtari2023gemtrans, amadou2024echoapex} first introduced by~\cite{arnab2021vivit}, where the frame encoder is a vision transformer (ViT) which produces frame-wise class token encodings that are subsequently fed through a temporal transformer. A final classification head is then attached to the temporal transformer's class token, and the model trainable for classification tasks. Full video transformers~\cite{zhang2024echo} \cite{szijarto2024masked} processing entire echo video clips have also been proposed, however this comes at computational cost due to the quadratic complexity of the self attention mechanism used in transformer blocks. To improve efficiency, works from~\cite{szijarto2024masked, yang2025echocardmae} use ultrasound triangle region of interest (ROI) masking in their video models to reduce the set of input tokens to only those localized to the ultrasound triangle.
\\
\indent
Training transformer encoders from scratch however is challenging, as the models generally require significantly more labeled training data than their CNN counterparts to achieve sufficient performance. Self supervised learning (SSL) has emerged as a powerful tool to remedy this problem, where a pre-text task is constructed using only unlabeled data with the intention to train the model to learn useful representations of the data without the need for ground truth annotations. Pre-trained models can then either be used as a starting point for tuning on downstream tasks, or alternatively the representations themselves used as inputs to lightweight classifiers. In imaging applications, examples of pre-text tasks include masked image modeling~\cite{he2022masked}, where input images are masked and the model tasked with reconstructing the original image from only the unmasked content. SSL methods have been successfully applied to echo~\cite{szijarto2024masked, zhang2024echo}, enabling models such as EchoApex~\cite{amadou2024echoapex} to be used as general backbones adaptable to multiple different tasks such as segmentation, view classification, and EF regression. However, whilst pure image and video deep learning models including both CNNs and transformers have achieved impressive diagnostic performance in the literature~\cite{akerman2023automated}, they are ``black boxes'' and give no assurance that their predictions corresponds to image content which relates to the pathology or anatomy in the image.
\subsection{Clinically Interpretable Models}
An alternative approach towards more interpretable models is to instead apply classifiers to clinically useful features extracted from images, videos or other relevant patient information. Examples from echocardiography have used measurements such as cardiac chamber size and areas extracted from segmentation maps as input to a CNN classifier~\cite{chiou2021ai}. Related works from~\cite{upton2022automated} and~\cite{sanchez2018machine} used classical machine learning models to process a plethora of hand crafted features calculated from contour points and myocardial strain curves, respectively. Whilst the classifier itself is still a ``black box'', it can at least only use features which a clinician would also use in making a diagnosis. 
\\
\indent
However, completely relying on features based on strain, motion or segmentation masks removes the ability of models to leverage texture information from the images or videos. In the clinical diagnosis of cardiac amyloidosis for example, a ``sparkling, hyper-refractile texture of the myocardium'' is mentioned as an abnormal parameter in CA practice guidelines~\cite{dorbala2021asnc} which would not be captured by models omitting pixel information from their input. Work from~\cite{hathaway2022ultrasonic, jamthikar2025ultrasonic} used hand crafted features derived from myocardium texture localized to the myocardium as classifier inputs, leveraging both structural and texture information. In our work, we show that by sampling image patches at myocardium point locations both anatomical pixel information and myocardium geometry can be embedded into input tokens for a transformer model, without the need for hand crafted texture features. These models retain the guarantee that predictions were made using clinically relevant features whilst also enabling the use of potentially valuable pixel information located in these regions. We note that a number of other works have embedded clinical knowledge and constraints into transformers, with~\cite{mokhtari2023gemtrans} encouraging attention weights to align with clinically relevant image regions through an additional loss, and~\cite{szijarto2024masked, yang2025echocardmae} utilizing ultrasound ROI masking to constrain transformers to the ultrasound triangle. Whilst these works make steps towards incorporating clinical domain knowledge into the transformer, neither are explicitly constrained to anatomical regions as in our prior work~\cite{thorley2025anatomically} and~\cite{hathaway2022ultrasonic, jamthikar2025ultrasonic}.
\subsection{Segmentation and Motion Tracking}
The majority general purpose pre-trained transformers in echocardiography have 
approached the problem of segmenting the myocardium and other cardiac structures as a semantic segmentation task~\cite{amadou2024echoapex, echofm}. Similarly, works adapting the foundation scale segment anything model (SAM)~\cite{kirillov2023segment}, trained on vast quantities of labeled natural images, have also been tuned semantic segmentation in echo~\cite{ravishankar2023sonosam, ravishankar2023sonosamtrack, deng2024memsam}. Pixel-wise segmentation masks however require an additional step to extract either a contour or mesh of points on each frame in order to calculate clinical metrics. Whilst alternative CNN approaches directly regressing point coordinates or B-Spline surfaces of the chamber wall contours have been proposed~\cite{porumb2021site, chernyshov2024automated, akbari2024beas}, applying these methods frame by frame does not strictly establish correspondence between points through time and may face difficulties capturing regional deformations.
\\
\indent
Establishing correspondences between pixels or points is a fundamental problem in computer vision and has been studied under a number of different guises including optical flow and deformable image registration. In echo, the problem has traditionally been tackled via speckled tracking where blocks of pixel intensity are tracked across frames by maximizing their similarity between frames~\cite{d2016two}. This assumes that the speckle pattern is consistent through time, which in practice does not always hold due to speckle decorrelation arising from imaging artifacts and through plane motion. Commercial systems rely on a combination of speckle tracking with specialised regularisation or shape priors to ensure the consistency of tracking, the specific details of which are industrial secrets~\cite{d2016two}. Advances in deep learning have enabled the minimization of popular image registration energy functionals via neural networks~\cite{wei2020temporal}. These methods can leverage auxiliary information like segmentation masks within their loss functions to assist with motion estimation in challenging areas of the image where there is little speckle pattern to track.
\\
\indent
Recently, the problem has been tackled as a supervised regression task using ``ground truth'' tracked points from simulated ultrasound sequences. These points can be interpolated to dense displacements to train supervised optical flow networks~\cite{evain2022motion,ostvik2021myocardial}. Alternatively, tracked points from speckle tracking systems validated and corrected by humans have been used as ``ground truth'' trajectories to train point tracking models~\cite{abulkalam2024echotracker,chernyshov2025low}. Clinical metrics can then be calculated directly from predicted point trajectories. All supervised methods are of course bounded in performance by the points upon which they are trained, which may themselves contain inaccuracies or flaws. However, they do not rely on any hand crafted similarity metrics or regularizers and can in theory be trained on a wide variety of both simulated and human validated tracked points. In our work, we move beyond the limitations of semantic segmentation used by other general-purpose transformers in echo~\cite{amadou2024echoapex, echofm} and show our ViACT can be used as a backbone for myocardium point tracking. We show that through our proposed anatomical MAE pre-training, we can remove the need for specialist model components such as correlation volumes used in state-of-the-art trackers~\cite{abulkalam2024echotracker, karaev2024cotracker, chernyshov2024automated}  and retain the capacity of model to function as a general backbone for a variety of echo analysis tasks.  
\section{Methods}
In this section, we present the ViACT architecture and corresponding anatomical MAE pre-training framework. We detail how the ViACT can be used as a general backbone for echocardiogram analysis, tuning models for regressing EF and point trajectories. We also show that the model can be tuned to classify cardiac disease localized to the anatomical structure used to constrain the model. In this work, we chose CA, which has many known imaging features localized to the myocardium~\cite{dorbala2021asnc}.
\subsection{Anatomically Constrained Transformers}\label{subsec:anatomically_constrained_transformers}
\begin{figure}[!t]
\centerline{\includegraphics[width=\columnwidth]{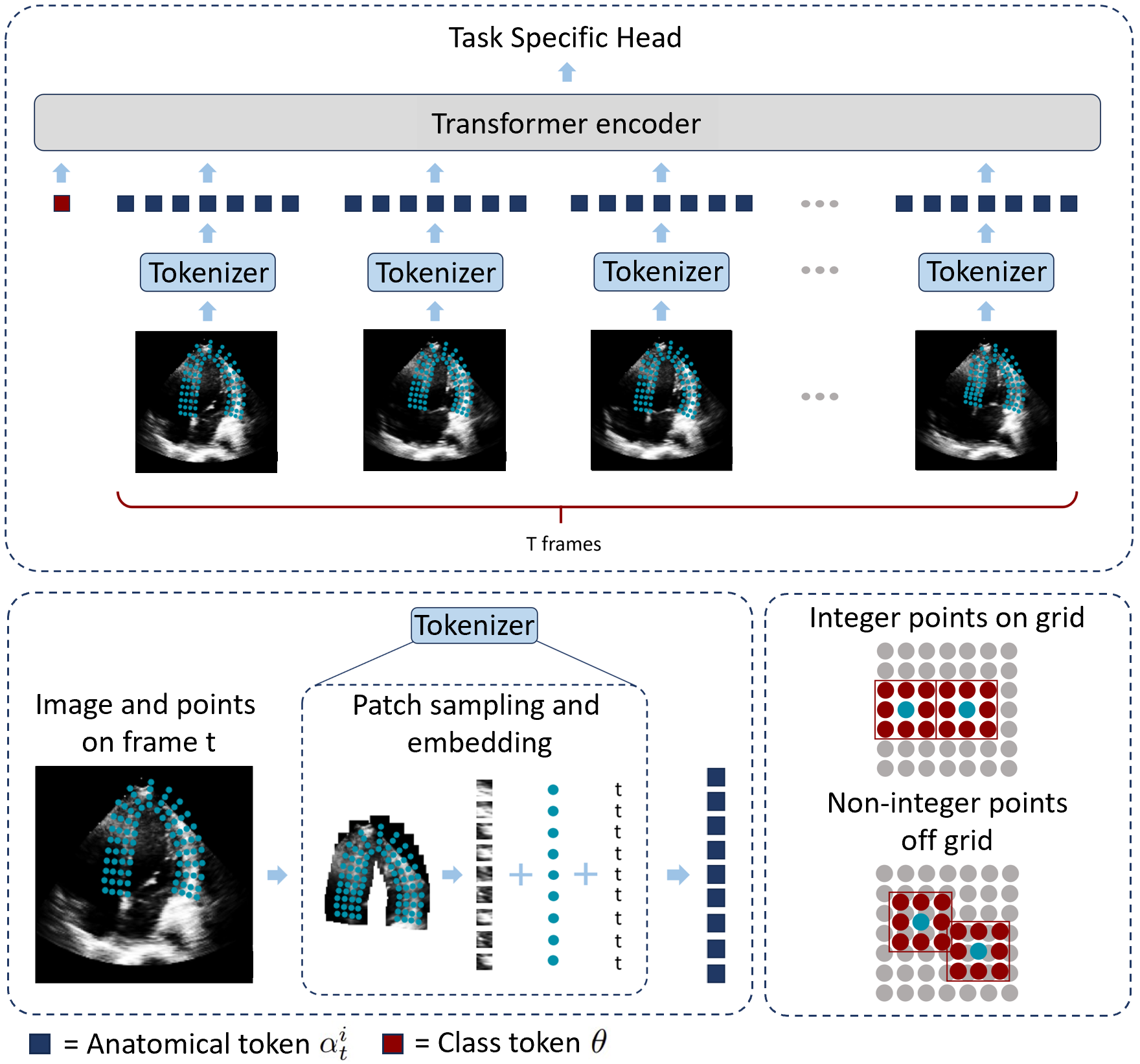}}
\caption{Top: the ViACT model. Bottom left: the tokenizer component of the model embedding a single frame and corresponding points. Bottom right: example $3 \times 3$ patches (red) centered
at integer and non-integer points (teal) on a grid of pixels (gray).}
\label{fig:temporal_viact}
\end{figure}
To introduce the ViACT model, we first assume that we have ultrasound video sequences $V_T = (I_0, I_1, ... , I_T)$ comprised of frames $I_t \in {\mathbb{R}^{H W}}$, where $t \in [0, T], T \in \mathbb{N}$. Our model leverages anatomical point trajectories in both the model and pre-training objective itself, which we constrain to only the myocardium for this work.
Formally, we assume that for each video $V_T$, we have obtained a sequence of points $P_T^N=(P_t^i)_{i=0, t=0}^{N, T}$, where $P_t^i = (x_t^i, y_t^i) \in {\mathbb{R}^{2}}$ represents the $i^{th}$ point of a myocardium point set on the $t^{th}$ frame, with $i \in [0, N], N \in \mathbb{N}$. These points must be obtained in advance by another model, simulation, or human annotation. We detail how we acquired points for our experiments in section~\ref{subsec:datasets}. We now present the ViACT model, with a summary graphic found in Fig~\ref{fig:temporal_viact}.
\\
\indent
\textbf{Tokenizer:} image and video transformers generally partition a video clip into spatial~\cite{he2022masked} or spatiotemporal~\cite{feichtenhofer2022masked} patches or tubes. This partitioning is centered at pixel locations on an integer grid allowing neighboring pixels to be accessed and extracted, often via a convolution. The myocardium point sets processed by our model deform through the video sequences, covering non-integer locations in each frame. We therefore cannot simply extract raw image pixels from the integer pixel grid as the myocardium points may lie between pixels. Instead, we propose to interpolate the original image pixel grid by creating a $j \times j$ sampling grid for each point. Each grid is centered on point $P_t^i$, where $j \in \mathbb{N}$ is the desired patch size and the grid spacing equal to a single pixel.
Fig~\ref{fig:temporal_viact}, bottom right, shows an example integer patch and non integer patch sampling. As the myocardium points deform, it is possible that there is an overlap between the sampled patches since points move nearer together. Following the construction of sampling grids, each frame $I_t$ is sampled with bilinear interpolation at the grid locations to produce a set of patches $C_T^N = (C_t^i)_{i=0,t=0}^{N,T}$, where $C_t^i \in {\mathbb{R}^{j j}}$. We then flatten each of these patches and linearly embed them to the transformer's embedding dimension of length $k$, where $k \in \mathbb{N}$. 
\\
\indent
We experimented with four different positional embeddings for our tokens, all leveraging the spatial coordinates of the tokens themselves. For the first, we embedded the original point coordinates of each patch as $\rho(P_t^i)$, where $\rho$ is either the summed \emph{sin cos} embedding of the $x$ and $y$ coordinates or a linear projection of point coordinates to a vector of length $k$. We refer to this positional embedding as a \emph{point} embedding in our experiments, with a \emph{sin cos} or \emph{linear} suffix depending on the method of embedding. We also experimented with embedding points relative to the apex point on the first frame in the sequence as $\rho(P_t^i - P_0^{apex})$, which we refer to as an \emph{apex relative} embedding henceforth. In contrast to our preliminary work~\cite{thorley2025anatomically}, this ensures that apical motion is also captured by the positional embeddings, whilst remaining invariant to the spatial location of the heart in the video. All four variants are evaluated in experiments of section~\ref{sec:Experiments}. A learnable temporal positional embedding for each frame in the clip is also added to encode the temporal position of the patch in the video. Patch and positional embeddings for all points and frames were then summed to produce a set of input tokens $\alpha_T^N=(\alpha_t^i)_{i=0, t = 0}^{N, T}$ for the transformer, where $\alpha_t^i \in {\mathbb{R}^{k}}$. A graphic showing the tokenizer can be seen in Fig~\ref{fig:temporal_viact}, bottom left.
\\
\indent
\textbf{Transformer:} the transformer component of the temporal ViACT model is a standard transformer encoder~\cite{vaswani2017attention}, processing all tokens $\alpha_T^N$ with an optional appended class token $\theta \in \mathbb{R}^k$. The transformer is applied to the input tokens to produce encodings of $\hat{\alpha}_T^N$ and $\hat{\theta}$ of length $k$, corresponding to the anatomical and class tokens, respectively. Task specific heads can then be attached to any of these encodings and the entire model tuned for a given task. Following~\cite{vaswani2017attention}, each transformer block in $\tau$ is composed of multi head self-attention followed by a multi-layer perceptron (MLP). Layer normalisation is applied before each, and a residual connection after. A graphic depicting the temporal ViACT model is shown in Fig~\ref{fig:temporal_viact}, top, for clarity. In contrast to our preliminary experiments~\cite{thorley2025anatomically}, there is no space-time factorization and each token can attend to any other token across the sequence allowing information to flow between all tokens as they deform through the video clip.
\subsection{Anatomical MAE}
\label{subsec:temporal_anatomical_mae}
\begin{figure}[!t]
\centerline{\includegraphics[width=\columnwidth]{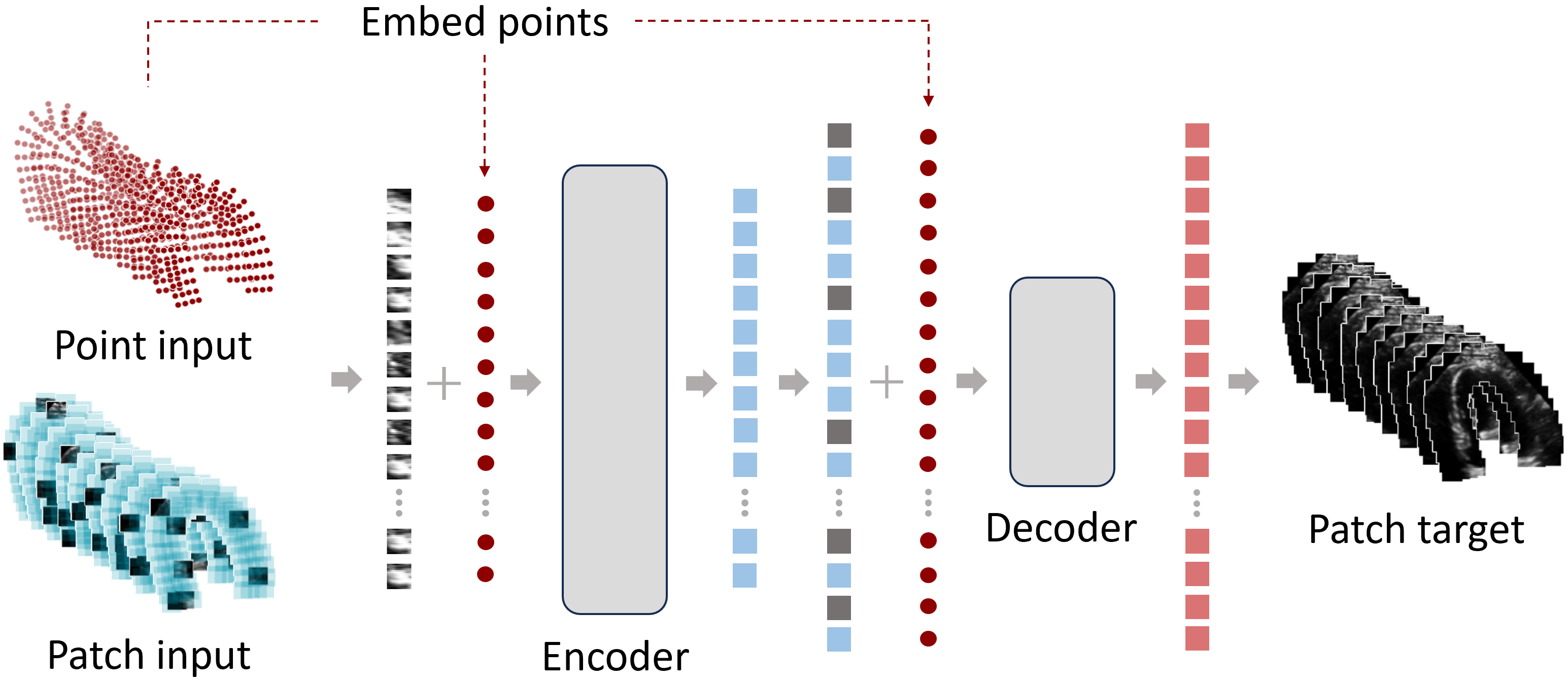}}
\caption{The anatomical MAE framework. Graphic inspired by \cite{he2022masked}.}
\label{fig:anatomical_mae}
\end{figure}
\begin{figure*}[ht] 
\centering  
{\includegraphics[width=\textwidth]{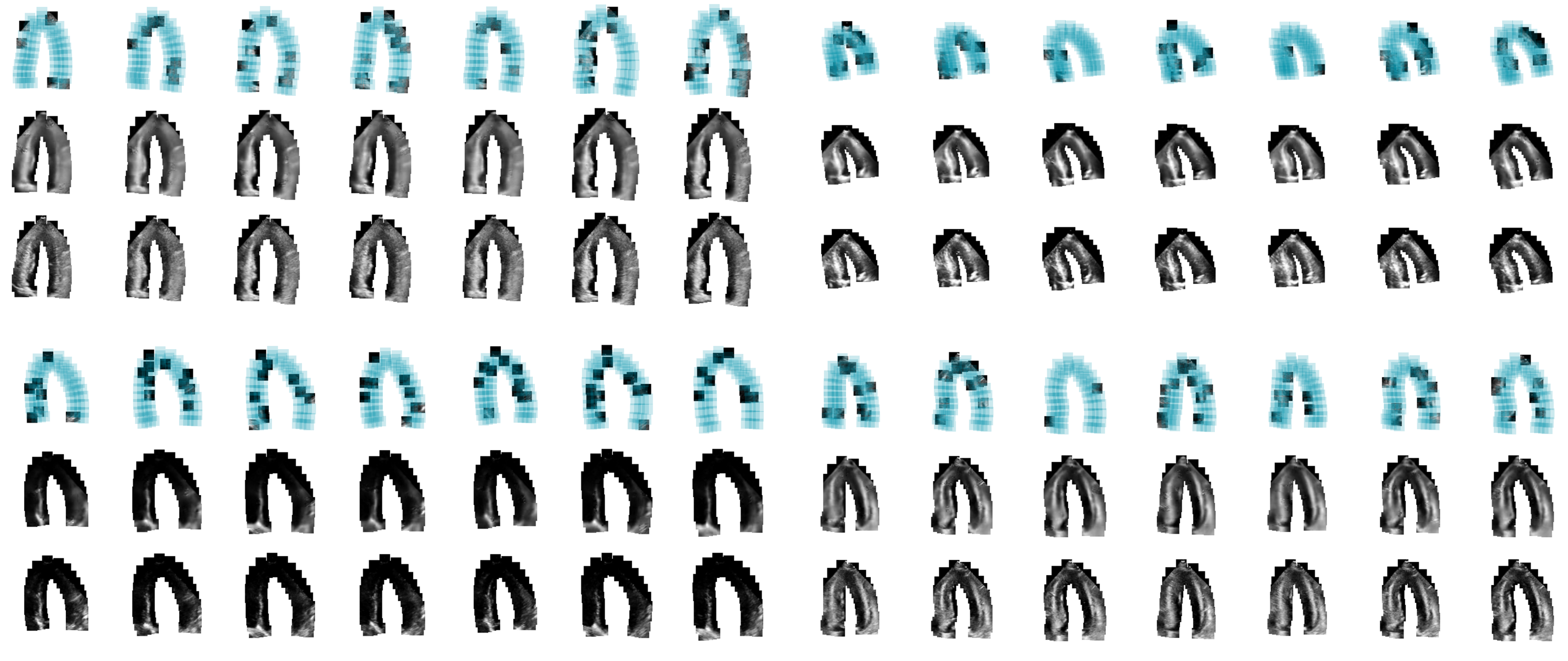}}
\caption{Example reconstructions from the temporal ViACT. For each sample, top row depicts masked patches, middle row reconstructed patches and bottom row ground truth patches from a subset of the 18 processed frames. Masked patches are depicted with transparent teal squares.}
\label{fig:reconstructions}
\end{figure*}
Work from~\cite{thorley2025anatomically} introduced an anatomical variant of the well known MAE framework~\cite{he2022masked} to pre-train an anatomically constrained transformer on sampled patches and corresponding myocardium point sets from individual 2D image frames. In a similar fashion to extensions of MAE to video data~\cite{feichtenhofer2022masked,tong2022videomae}, we extend the anatomical MAE framework to pre-train our temporal ViACT model. Mirroring its 2D anatomical MAE counterpart from~\cite{thorley2025anatomically}, our temporal version pairs the ViACT encoder with a smaller decoder transformer. For each video sequence $V_{T}$, patches are extracted from each frame at the myocardium point locations. A portion of these are masked and the remaining unmasked tokens are embedded via our tokenizer and fed through the ViACT encoder. The output tokens from the encoder are then padded with mask tokens to the same order as the original tokens, and embeddings of the original myocardium point locations along with a learnable temporal positional embedding are added to the full set of tokens. This set of tokens is passed through the smaller decoder transformer, with the output tokens then passed through a reconstruction head. A mean squared error loss between the reconstructed patches corresponding to the masked tokens and the original image patches is then used to train the network. By only masking and reconstructing anatomical tokens instead of the full video sequences, we greatly reduce the number of tokens to be processed by the model. This is of great importance to the compute requirements of the pipeline when using video data as attention is quadratic in complexity with regards to the number of tokens. A graphic depicting the temporal anatomical MAE framework is shown in Fig~\ref{fig:anatomical_mae}, with example reconstructions in Fig~\ref{fig:reconstructions}.
\subsection{Task Specific Tuning}
In contrast to work from~\cite{thorley2025anatomically} where a pre-trained ViACT was tuned for only a classification task, we show that our temporal ViACT model can also be used for point tracking and EF regression. This enables us to pre-train a single model that is adaptable to common echo processing tasks. We detail task specific tuning for the three tasks below.
\\
\indent
\textbf{Point tracking:} in this work we formulate myocardium point tracking as a fully supervised point tracking task similarly to~\cite{abulkalam2024echotracker, chernyshov2025low}. We assume we have a set of ground truth tracked points $P_T^N$ along the myocardium for each echo video $V_T$ to be used as a regression target. These points may be obtained via synthetic simulations as in~\cite{evain2022motion,ostvik2021myocardial}, or may be human verified (and if necessary, corrected) points from a speckle tracking system as in~\cite{abulkalam2024echotracker, chernyshov2025low}. The goal of a point tracking network is to track a set of query points (in our case, the myocardium) from the first frame of a video clip through the entire sequence so they align with the ground truth trajectories. Popular works~\cite{karaev2024cotracker, harley2022particle} first initialize a copy of the query points on each frame in the video, which they deform to align with the desired trajectory. This style of point tracker has seen recent success tracking LV and RV point sets in echos~\cite{abulkalam2024echotracker, chernyshov2025low}. Whether utilizing a transformer~\cite{chernyshov2025low} or CNN~\cite{abulkalam2024echotracker} backbone, the models utilize specialized components from optical flow literature such as correlation volumes~\cite{teed2020raft} as integral components to the models. We forego these components and show that anatomical MAE is a sufficiently powerful pre-text task for a pure transformer model to learn correspondences between myocardium patches in a self supervised fashion instead.
\\
\indent
Formally, let us assume as in other point tracking works~\cite{abulkalam2024echotracker, karaev2024cotracker, chernyshov2025low, harley2022particle} that we have obtained a set of myocardium query points $P_0^N$ for the first frame of the sequence which we use to initialize tracking point on each frame as $\tilde{P}_T^N = ( \tilde{P}_t^i)_{i=0,t=0}^{N,T}$ where $\tilde{P}_t^i = {P}_0^i ~ \forall ~ t \in T$. As per the ViACT model described in section~\ref{subsec:anatomically_constrained_transformers} we use these points to sample image patches which are embedded along with their corresponding points into tokens $\alpha_T^N$. A class token is not necessary for the tracking task, but as the ViACT model is intended as a general backbone it is pre-trained with a class token $\theta$ for use in other downstream tasks. There is no linear head or loss acting upon it when tuning for the tracking task however. The input tokens are then processed by the pre-trained ViACT transformer $\tau$ producing output encodings $\hat{\alpha}_T^N$. Each of these encodings are passed through a linear layer mapping from the embedding dimension $k$ to $\Delta \tilde{P}_T^N = (\Delta \tilde{P}_t^i)_{i=0,t=0}^{N,T}$, where $\Delta \tilde{P}_t^i = (\Delta \tilde{x}_t^i, \Delta \tilde{y}_t^i) \in {\mathbb{R}^{2}}$ are displacements used to deform the initialized points $\tilde{P}_T^N$. Similarly to other recent point tracking works~\cite{abulkalam2024echotracker,karaev2024cotracker,chernyshov2025low}, the model can then be trained under a supervised loss- we chose the sum of absolute differences between the ground truth and deformed points as:
\begin{equation} L_{diff} =  |P_T^N - (\tilde{P}_T^N + \Delta \tilde{P}_T^N)|\end{equation}\label{eq:sad_loss}
\begin{figure}[!t]
\centerline{\includegraphics[width=\columnwidth]{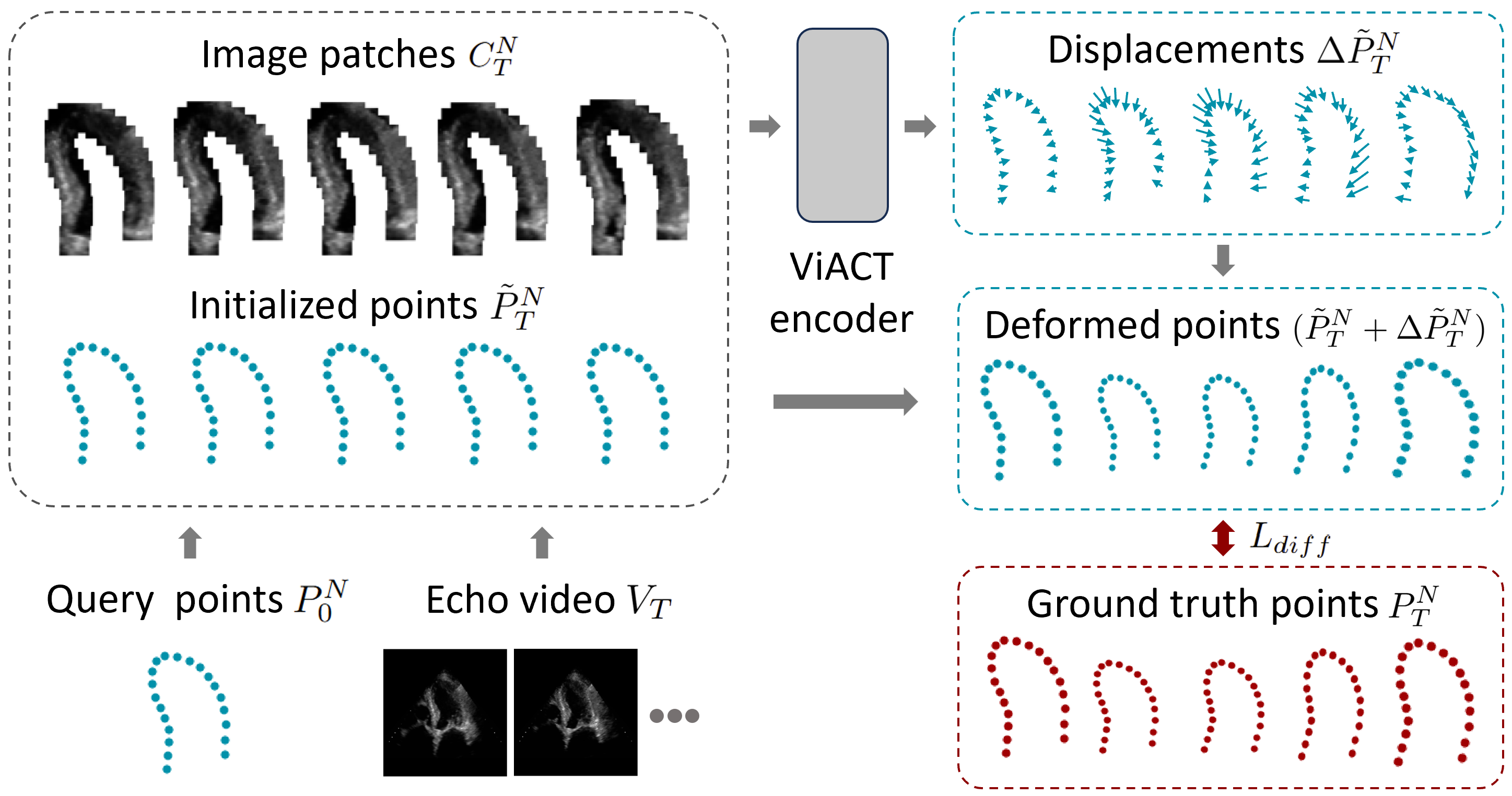}}
\caption{The pipeline for tuning a ViACT model for the point tracking task.}
\label{fig:viact_tracking}
\end{figure}
A graphic summarizing the point tracking pipeline on a small number of frames and points can be found in Fig~\ref{fig:viact_tracking} for reference. As the displacement of the myocardium is relatively small, at least within the 18 frame sequences used in our experiments, we found that the $16\times16$ sized patches provided sufficient information for the model to determine accurate trajectories of points in a single pass. However, in the case that expected trajectories move far beyond the patch size, the model can in theory be applied iteratively similarly to~\cite{karaev2024cotracker} to further refine the point trajectories by resampling input patches with the deformed points from each pass. The ViACT model assumes no order, structure or number of myocardium points as the positional embedding of each token is constructed from the point coordinates, with the exception of requiring a single apex point in the case of the \emph{apex relative} variant. We can therefore use a ViACT model pre-trained on a different number of myocardium points to that which we wish to track.
\\
\indent
\textbf{Disease classification:} A linear classification head can be attached to the class token encoding $\hat{\theta}$ from a pre-trained ViACT and the entire model tuned for a given classification task under a binary cross entropy loss. We extensively evaluate the performance of the video ViACT tuned for a CA classification task in section \ref{sec:Experiments}. Similarly to~\cite{thorley2025anatomically}, the model assumes that myocardium point trajectories have already been obtained to constrain the model for classification tasks. In theory, this facilitates a human-in-the-loop to verify the quality of the myocardium point trajectories and thus ensure the model is focused on the myocardium. 
\\
\indent
\textbf{EF regression:} Quantification of left ventricular (LV) cardiac function in clinical practice is typically derived from 2D echocardiography. End-systolic (ES) and end-diastolic (ED) volumes are used to calculate EF, a marker of overall systolic contraction. The volumes are commonly approximated with the modified Simpsons biplane method~\cite{lang2015recommendations}, a stack of disks controlled by epicardium contours at ED and ES from both two and four chamber views. In practice, the four and two chamber views from which contours are extracted are often foreshortened, which can introduce variability between measurements. With the rise of deep learning EF can instead be regressed using ``ground truth'' EF values obtained from 2D or 3D echocardiography, or even CMR imaging. To adapt our model for EF regression, we simply attach a regression head to the class token in a similar fashion to the CA classification task. We used a linear head mapping to a single value, passed through a sigmoid to predict $EF_{pred}$. Given ground truth values $EF_{gt}$, the model can be trained under the mean squared error between $EF_{pred}$ and $EF_{gt}$.
\section{Experiments and Results}\label{sec:Experiments}
In this section, we present our experimental results for the three echo processing tasks. We provide key implementation details including model hyperparameters, pre-training and tuning setups, as well as information on our datasets and pre-processing steps. We then present our experiments on the point tracking, CA classification and EF regression tasks.
\subsection{Implementation Details} 
\label{subsec:Implementation_details}
Unless otherwise stated, we used a ``tiny'' scale transformer for all our experiments. Specifically, the embedding dimension $k$ of each transformer block was set to 192 with 3 heads. We used a total of 12 transformer blocks, with an MLP hidden dimension of 768 and a GeLU activation function. We trained and tuned our models using an AdamW
optimizer~\cite{loshchilov2016sgdr} with weight decay of 0.05 and momentum values of (0.9, 0.999). Batch sizes and base learning rates for each specific task are provided in their corresponding subsection. We scaled each base learning rate by batch size / 256 following~\cite{goyal2017accurate} to determine the final base learning rate. We ran all of our experiments on two NVIDIA GeForce RTX 3090 24.5G GPUs, and implemented the models in PyTorch 3.1.
\subsection{Datasets}
\label{subsec:datasets}
To train and evaluate our model for the three tasks, we used a combination of one private and two public datasets depending on the task and availability of adequate labeling. The datasets and any pre-processing are detailed below, with each following subsections detailing when and how the datasets were used for our experiments.
\\
\indent
\textbf{Private CA dataset:} following preliminary work from~\cite{thorley2025anatomically}, we used a private dataset of 1959 4-chamber echocardiograms. The dataset contained 1520 controls and 439 CA cases which were a mix of both \emph{amyloid transthyretin} (ATTR) and \emph{amyloid light-chain} (AL) patients. The CA cases met the criteria detailed in~\cite{Sharmila} for the clinical diagnosis of CA, and controls were comprised of both healthy patients and patients with other pathologies as described in~\cite{thorley2025anatomically}. Patients were retrospectively selected from a variety of collaborators participating in a study on CA by the University of Chicago. Ethical approval was obtained by the respective institutional review boards of each collaborator. The CA classification task was chosen as the pathology is known to impact left ventricle structure and function where our model is explicitly focused, with patients exhibiting increased LV wall thickness and apical sparing in regional strain patterns~\cite{dorbala2021asnc}. The dataset also contained manually acquired EF values for each patient, with a mean of $58.9 \pm 10.9 \%$ and range of $[10.9\%, 85.0\%]$, enabling us to train an EF regressor upon the same dataset. For each of the DICOM echocardiograms, we used the same pre-processing steps as presented in~\cite{thorley2025anatomically}. Specifically, we grayscaled all image frames, removed content outside of the ultrasound region, resized frames to $224\times224$ pixels and resampled the sequence of frames to a constant frame time of 33.33ms. For each clip, we obtained 21 myocardium contour points with a point regression U-Net~\cite{porumb2021site} used in~\cite{asch2022human}, adding an additional three perpendicular rows of with an interval of six pixels. We used the same 70/15/15 split as~\cite{thorley2025anatomically} for training, validation and testing sets which resulted in 1372, 294 and 293 unique patients, respectively. The prevalence of CA cases was approximately the same in all three sets. We used these train/val/test splits for both the CA classification and EF regression tasks, and used the training set data for pre-training our models.
\\
\indent
\textbf{CAMUS synthetic:} the CAMUS synthetic dataset from Evain et al. comprises of 100 simulated A4C echo videos with corresponding tracked points~\cite{evain2022motion}. We removed two patients from the dataset which we observed to have poor tracking. Each video was available with and without synthetic reverberation artifacts, resulting in a total of 196 video sequences and points comprised of varying numbers of frames. Each clip and corresponding set of points was resized to $224\times224$ pixels.
\\
\indent
\textbf{Leuven simulated:} we used a second synthetic dataset consisting of a simulated ultrasound sequences for 7 different commercial Ultrasound vendors from~\cite{alessandrini2017realistic}. For each vendor, a 4, 3 and 2 chamber simulation was provided for a healthy patient and 4 abnormal patients. This resulted in a total of 105 clips. Similarly to the CAMUS synthetic dataset, we resized each clip and points to $224\times224$ pixels. Details of how we used this dataset in conjunction with CAMUS synthetic are found in section \ref{subsec:tracking_results}.
\subsection{Pre-training}
\label{subsec:pre_training}
\begin{figure}[!t]
\centerline{\includegraphics[width=\columnwidth]{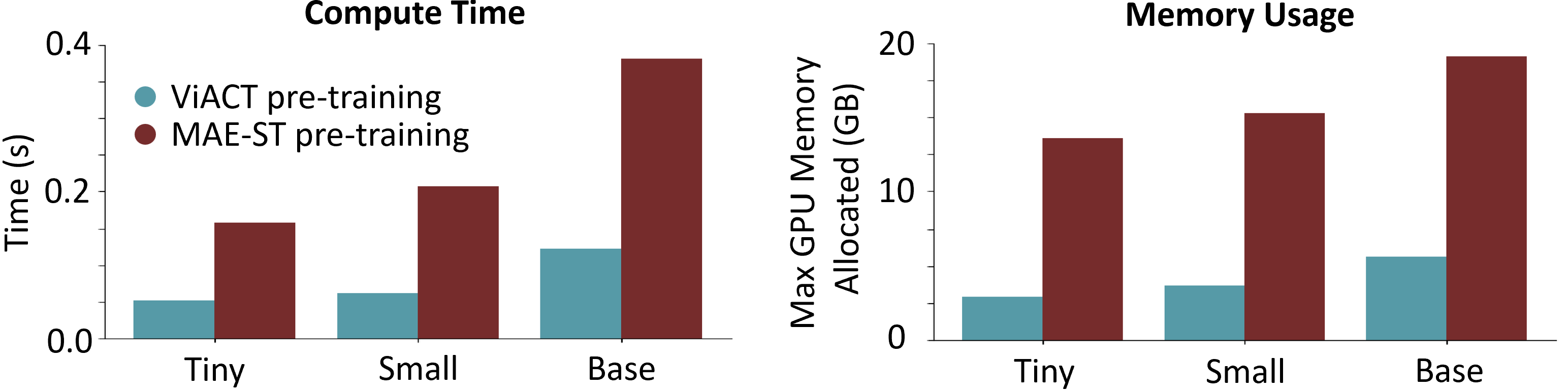}}
\caption{Pre-training compute time and memory usage for a ViACT and MAE-ST~\cite{feichtenhofer2022masked} with varying model size, using a batch size of 10.}
\label{fig:compute_experiments}
\end{figure}

For pre-training, we utilized all full length clips and points from the training set of our private CA dataset. For each epoch, we sampled 40000 frame and corresponding point sequences covering 18 frames with random starting frames and spacing. We pre-trained our ViACT model under the anatomical MAE scheme described in section~\ref{subsec:temporal_anatomical_mae} using a linear warm up of 200 epochs followed with cosine decay~\cite{loshchilov2016sgdr} for 1800 epochs. We used a batch size of 160 with a base learning rate of 1.5e-4. To determine the optimal masking ratio (the proportion of tokens masked in the anatomical MAE pre-training), we pre-trained \emph{point linear} ViACTs with masking ratios 
between [0.8, 0.95] at increments of 0.05. In line with work from~\cite{feichtenhofer2022masked} on spatiotemporal MAE, we found a masking ratio of 0.9 produced optimal results on the CA classification validation set. We proceeded to use a masking ratio of 0.9 for all subsequent ablation studies and experiments. We used a tiny scale model for our experiments, but show in Fig~\ref{fig:compute_experiments} that our framework is much lighter than an equivalent video transformer pre-training with MAE-ST~\cite{feichtenhofer2022masked} across model scales due to the reduction in input tokens.
\begin{table*}[ht!]
        \caption{Ablation study on validation sets. Accuracy and weighted F1 score are for CA classification, and ME is for point tracking.}
	\centering
	{
		\begin{tabular}{|c|c|c|c|c|c|}
			\hline
			\textbf{Positional embedding} & \textbf{Factorized} & \textbf{Pre-trained} & \textbf{Accuracy} $\uparrow$ & \textbf{Weighted F1} $\uparrow$ & \textbf{ME (pix)} $\downarrow$ \\ \hline
Apex relative sin cos & $\times$  & $\times$ & 80.41 $\pm$ 1.83 & 77.76 $\pm$ 3.15 & 1.78 $\pm$ 1.64
\\
" & $\times$  & \checkmark & 81.79 $\pm$ 1.43 & 80.03 $\pm$ 2.04 & \textbf{0.74 $\pm$ 0.79}
\\
" & \checkmark  & $\times$ & 79.89 $\pm$ 1.49 & 77.90 $\pm$ 3.77 & 1.70 $\pm$ 1.50
\\
" & \checkmark  & \checkmark & 79.54 $\pm$ 1.49 & 75.61 $\pm$ 4.38 & 1.77 $\pm$ 1.44
\\
\hline
Apex relative linear & $\times$  & $\times$ & 81.14 $\pm$ 1.65 & 79.88 $\pm$ 3.19 & 1.75 $\pm$ 1.65
\\
" & $\times$  & \checkmark & 81.93 $\pm$ 1.59 & 80.14 $\pm$ 2.54 & 0.79 $\pm$ 0.80
\\
" & \checkmark  & $\times$ & 81.04 $\pm$ 2.00 & 79.35 $\pm$ 3.52 & 1.78 $\pm$ 1.58
\\
" & \checkmark  & \checkmark & 80.26 $\pm$ 1.64 &  76.29 $\pm$ 4.11 & 1.73 $\pm$ 1.57
\\
\hline
Point sin cos & $\times$  & $\times$ & 79.63 $\pm$ 1.68 & 76.33 $\pm$ 3.19 & 1.97 $\pm$ 1.83
\\
" & $\times$  & \checkmark & 82.71 $\pm$ 1.65 & 81.30 $\pm$ 2.29 & 0.95 $\pm$ 1.01
\\
" & \checkmark  & $\times$ & 79.50 $\pm$ 1.57 & 76.49 $\pm$ 3.77 & 2.00 $\pm$ 1.79
\\
" & \checkmark  & \checkmark & 81.65 $\pm$ 1.06 & 79.74 $\pm$ 2.01 & 1.78 $\pm$ 1.56
\\
\hline
Point linear & $\times$  & $\times$ & 81.22 $\pm$ 1.65 & 79.35 $\pm$ 2.01 & 1.80 $\pm$ 1.69

\\
" & $\times$  & \checkmark & 83.53 $\pm$ 1.94 & 82.35 $\pm$ 2.37 & 0.81 $\pm$ 0.84
\\
" & \checkmark  & $\times$ & 80.24 $\pm$ 1.41 & 78.17 $\pm$ 2.59 & 1.77 $\pm$ 1.56
\\
" & \checkmark  & \checkmark & \textbf{83.76 $\pm$ 1.25} & \textbf{82.62 $\pm$ 1.65} & 1.65 $\pm$ 1.51
\\ \hline
	\end{tabular}}\label{tab:ablation_study}
\end{table*}
\begin{table*}[ht]
\caption{ME (mm) $\downarrow$ on simulated apical four chamber echos from~\cite{alessandrini2017realistic} for all vendors.}
	\centering
	{
		\begin{tabular}{|c|c|c|c|c|c|c|c|}
			\hline
			\textbf{Model} & \textbf{GE} & \textbf{Siemens} & \textbf{Samsung} & \textbf{Philips} & \textbf{Hitachi} & \textbf{Esaote} & \textbf{Toshiba} \\ \hline
            CotrackerV2~\cite{karaev2024cotracker} & 2.11 $\pm$ 2.01& 2.46 $\pm$ 2.50& 1.77 $\pm$ 1.57& 1.82 $\pm$ 1.71& 1.73 $\pm$ 1.38& 1.96 $\pm$ 1.89& 1.94 $\pm$ 1.86
            \\EchoTracker~\cite{abulkalam2024echotracker} & 1.36 $\pm$ 1.50& 1.61 $\pm$ 1.65& 2.36 $\pm$ 2.60& \textbf{1.13 $\pm$ 1.03}& 1.52 $\pm$ 1.58& 1.55 $\pm$ 1.64& \textbf{1.01 $\pm$ 0.93}
			\\
            MyoTracker~\cite{chernyshov2025low} & 1.80 $\pm$ 1.47& 1.99 $\pm$ 1.86& 2.84 $\pm$ 2.48& 1.54 $\pm$ 1.29& 1.84 $\pm$ 1.52& 1.66 $\pm$ 1.34& 2.09 $\pm$ 1.69
                \\
            ViACT factorized~\cite{thorley2025anatomically} & 2.15 $\pm$ 1.52& 2.72 $\pm$ 2.43& 3.78 $\pm$ 2.80& 2.25 $\pm$ 1.89& 3.25 $\pm$ 2.70& 2.34 $\pm$ 1.83& 2.05 $\pm$ 1.81
                \\
            ViACT & \textbf{0.88 $\pm$ 0.73}& \textbf{1.10 $\pm$ 0.92}& \textbf{1.42 $\pm$ 1.04}& 1.27 $\pm$ 1.05& \textbf{1.17 $\pm$ 0.88}& \textbf{1.13 $\pm$ 1.00}& 1.07 $\pm$ 0.96
                \\ \hline
\end{tabular}}\label{tab:tracking_results}
\end{table*}
\subsection{Point Tracking}
\label{subsec:tracking_results}
In this section, we detail our experimental results using the ViACT model for the point tracking task. We used the two synthetic tracking datasets, taking the 36 myocardium centerline points for each patient. We used all 4 chamber clips for all vendors and patient types from the Leuven set as our test set, resulting in 35 unique clips. For our training and validation set, we combined all remaining clips across both datasets with an 80/20 split. We used the same pre-trained backbones as the regression/classification tasks for our experiments. Despite both our own and comparative models being capable of processing sequences longer than 18 frames, we trained all models with a sequence length of 18 for a direct comparison with our pre-trained backbones which were fixed to this length. For training and validation, we used 18 frame sequences sampled from each patient at random starting frames with random spacing to enable use to directly compare with our pre-trained backbones. On the test set, we tracked points from the first frame through each 18 frame block using the final tracked points from the previous block as the set of query points for the next. As shown in~\cite{chernyshov2025low}, we would expect the performance of all models to improve further if trained with a larger sequence length. We used the mean absolute error (ME) between the predicted and ground truth trajectories as our evaluation metric.
\\
\indent
\textbf{Ablation study:} for our ablation study, we tested our ViACT with and without pre-training using the four different positional embeddings. Results are presented in table~\ref{tab:ablation_study}, where we observed consistent results accross the various positional embedding variants, and  a reduction in MAE of around one pixel with the pre-trained ViACT. To compare with the factorized ViACT from~\cite{thorley2025anatomically}, we took pre-trained ViACT frame-encoders and added a temporal transformer to process all frame-wise patch tokens. We then attached a tracking head mirroring our ViACT tracker to the encoded patch tokens. Information could therefore only flow between tokens on different frames through the final temporal transformer. Compared with our pre-trained video ViACT, we found that pre-training the frame encoders did not result in equivalent improvements. This highlights the importance of spatiotemporal anatomical MAE for the tracking task, where temporal correspondences between patches are inherently of great importance.
\\
\indent
\textbf{Benchmarks:} On the test set, we compared with MyoTracker~\cite{chernyshov2025low}, a lightweight adaption of CoTracker~\cite{karaev2024cotracker} previously evaluated on RV tracking. We also compared with the full scale CoTrackerV2~\cite{karaev2024cotracker} which we trained from scratch, removing the visibility component of the model as we consider tracking points to always be visible in this work. Finally we compared with EchoTracker~\cite{abulkalam2024echotracker}, which has been shown to outperform a number of state-of-the-art point trackers including Cotracker. All methods including our own follow the paradigm of deforming an initialized sequence of points across a video sequence, and were all trained on sequences of 18 frames using batch sizes and learning rates reported in the original papers.
\\
\indent
\textbf{Results:} Experimental results on the test set of 4 chamber clips across all seven vendors can be found in Fig~\ref{tab:tracking_results}. In keeping with the philosophy of this work to use a single pre-trained backbone for all echo processing tasks, we used the \emph{point linear} variants of both ours and the factorized ViACT as these were most performant across the full spectrum of tasks in our ablation study. We found our ViACT exceeded the performance of comparative methods on the majority of vendors. The factorized ViACT performed noticeably worse than all other methods, which was not unexpected given the limited opportunity for the model to establish temporal correspondences through the temporal transformer attached to the frame encoders. 
Our results show that with anatomical MAE pre-training, we can remove the need for domain-specific components used in state-of-the-art trackers~\cite{chernyshov2025low, abulkalam2024echotracker}.
\subsection{CA Classification and EF Regression}
\label{subsec:amyloid_experiments}
In this section we detail our experiments on the CA classification and EF regression tasks. We used accuracy and a weighted F1 score as evaluation metrics for CA classification, with mean absolute error (ME) and root mean squared error (RMSE) for EF regression. We repeated experiments fifty times with random seeds, presenting mean and standard deviation for each metric.
\\
\indent
\textbf{Ablation study:} We tested all combinations of the positional embeddings described in section~\ref{subsec:anatomically_constrained_transformers} with and without the space-time factorization from~\cite{thorley2025anatomically} on the CA classification task. Results are presented for both pre-trained and randomly initialized models, where we note that only the frame encoder of the \emph{factorized} models undergo pre-training as per~\cite{thorley2025anatomically}. Results on the validation set are presented in table~\ref{tab:ablation_study}. We found pre-training to consistently improve both accuracy and F1 score for ViACTs with and without factorization. We found \emph{point linear} positional embeddings to be optimal, and observed a similar level of performance with and without factorization.
\\
\indent
\textbf{Benchmarks:} We compared our model with a variety of other popular models for video processing. Specifically, we compared with a \emph{MAE-ST} from~\cite{feichtenhofer2022masked}, where we used the same tiny scale transformer as in our own experiments with patch size of $1\times16\times16$. The model was pre-trained on the same dataset as our ViACT, however we had to use a reduced batch size of 30 as this was the maximum capacity of our compute. We also compared with a \emph{factorized ViViT}, where we pre-trained only the ViT frame encoder component with the original image based MAE framework~\cite{he2022masked}. We also included our \emph{factorized ViACT} and a ViVit constrained to the ultrasound triangle, referred to as \emph{ROI-ViViT}, from our preliminary work~\cite{thorley2025anatomically}.
\begin{table}[ht!]
    \caption{Model performance on test set for CA classification task.}
\centering
\fontsize{8}{10}\selectfont
{
\begin{tabular}{|l|l|l|l|}
\hline
\textbf{Model} & \textbf{Accuracy} $\uparrow$ & \textbf{Weighted F1} $\uparrow$ \\
\hline
ViViT~\cite{arnab2021vivit} & 78.08 $\pm$ 2.41 & 75.98 $\pm$ 3.69 \\
ROI-ViViT  & 78.03 $\pm$ 2.93 & 75.82 $\pm$ 3.73 \\
MAE-ST~\cite{feichtenhofer2022masked} & 80.25 $\pm$ 1.65 & 76.87 $\pm$ 3.24 \\
ViACT factorized~\cite{thorley2025anatomically} & 81.58 $\pm$ 1.70 & 80.69 $\pm$ 1.74
\\
ViACT & \textbf{81.84 $\pm$ 1.89} & \textbf{80.82 $\pm$ 2.16}
\\
\hline
\end{tabular}
}
    \label{tab:classification_results}
    \end{table}
\begin{table}[ht!]
    \caption{Model performance on test set for EF regression task.}
\centering
\fontsize{8}{10}\selectfont
{
\begin{tabular}{|l|l|l|l|}
\hline
\textbf{Model} & \textbf{ME} $\downarrow$ & \textbf{RMSE} $\downarrow$ \\
\hline
ViViT~\cite{arnab2021vivit} & 7.42 $\pm$ 0.19 & \textbf{9.44 $\pm$ 0.18} \\
MAE-ST~\cite{feichtenhofer2022masked} & 7.47 $\pm$ 0.09 & 9.48 $\pm$ 0.07 \\
ViACT factorized~\cite{thorley2025anatomically} & 7.46 $\pm$ 0.05 & 9.56 $\pm$ 0.05
\\
ViACT & \textbf{7.39 $\pm$ 0.10} & 9.47 $\pm$ 0.11
\\
\hline
\end{tabular}
}
    \label{tab:regression_results}
    \end{table}
\\
\indent
\textbf{Results:} Experimental results on the test set are presented in table~\ref{tab:classification_results} for the CA classification, and table~\ref{tab:regression_results} for EF regression. We see that both the factorized and video ViACT models out-perform all other models on the CA classification task. Despite the ViACT being localized to the LV, which includes the myocardium from which EF is calculated, we found the performance of our model to be similar to the other video transformers.
\\
\indent
\begin{figure*}[ht!] 
\centering  
{\includegraphics[width=\textwidth]{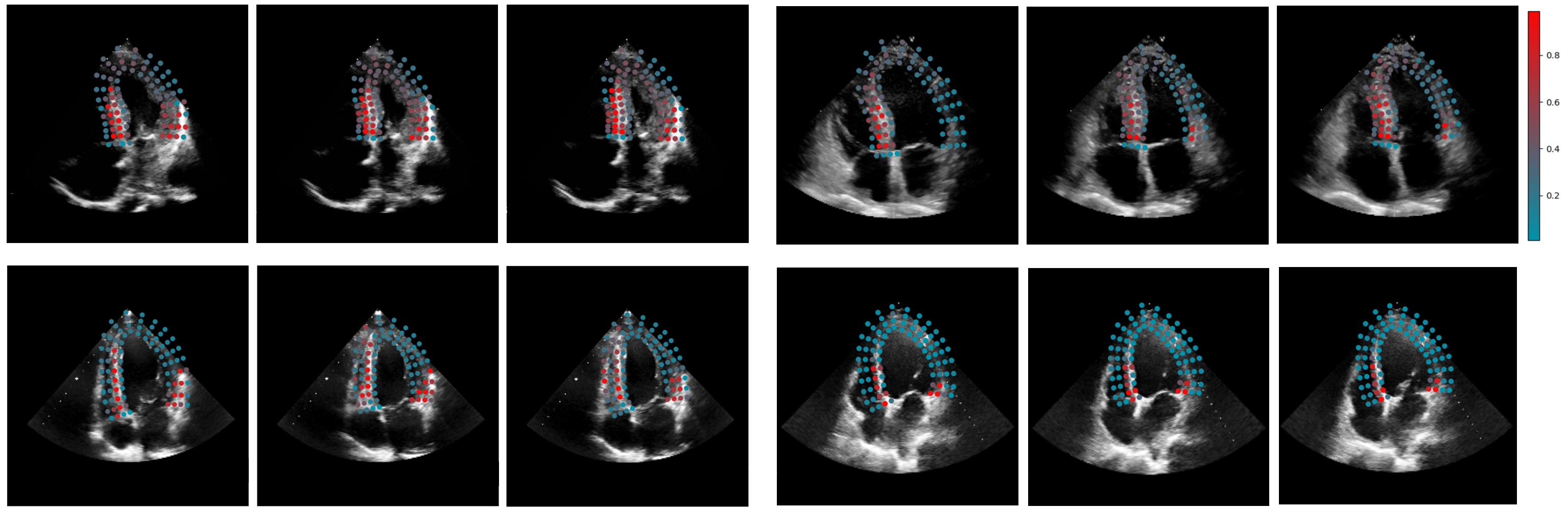}}
\caption{Attention maps for a selection of frames in a clip. Attention maps correspond to the class token attention scores from a single head of the final transformer block of the model. Myocardium points are colored with normalized attention values.}
\label{fig:attention_maps}
\end{figure*}
\textbf{Attention visualization:} the self-attention mechanism in transformers has been shown to focus on clinically relevant areas of echos such as valves in the case of aortic stenosis detection on select cases~\cite{amadou2024echoapex}. However, regular vision transformers offer no guarantee that attention is focused on relevant content in the image~\cite{thorley2025anatomically}. In contrast, we can visualize ViACT attention maps across the myocardium points from which patches were sampled. We calculated attention maps from a single transformer head in the final transformer block of the model. We took attention scores for the class token, reshaped the scores to align with the corresponding 18 frames $\times$ 84 points, and normalized across the entire sequence. Our attention maps are presented in Fig~\ref{fig:attention_maps} across a sample of three frames from each patient, with video examples found in the “Supplementary Files” tab on IEEE Author Portal.
\\
\indent
Qualitatively, we frequently found attention scores to cluster around the basal and mid sections of the myocardium as seen in Fig~\ref{fig:attention_maps}. This localization aligns with the known CA abnormality of apical sparing, a severe reduction in longitudinal strain across the basal and mid myocardium regions listed in patient care guidelines~\cite{dorbala2021asnc}.
\section{Conclusion}
In this paper, we introduced the ViACT, an anatomically constrained transformer and MAE pre-training strategy specifically designed for a variety of echo analysis tasks. The fine-tuned ViACT achieved high performance on CA classification and EF regression, and exceeded the performance of state-of-the-art point trackers across most vendors in our test set. By constraining tokens to only a set of myocardium points, we greatly reduced compute requirements of pre-training compared with full video models. Furthermore, the ViACT provides an explicit guarantee that any classification corresponds to the myocardium and enables the visualization of attention maps localized to the myocardium.
\\
\indent
A remaining limitation of our work is that a set of myocardium query points are still required in order to initialize point tracking. This is a limitation of all current point trackers in echo~\cite{abulkalam2024echotracker,chernyshov2024automated}, requiring either manual delineation of contour points or a further segmentation step to acquire the points independently. In future work we intend to incorporate this step within the ViACT framework. This would enable the end-to-end analysis of echo clips, both segmenting and tracking the myocardium to calculate clinical metrics as well detecting cardiomopathies such as CA from the tracked points and corresponding patches.  
\bibliographystyle{IEEEtran}
\bibliography{ref}

\begin{thebibliography}{10}
\providecommand{\url}[1]{#1}
\csname url@samestyle\endcsname
\providecommand{\newblock}{\relax}
\providecommand{\bibinfo}[2]{#2}
\providecommand{\BIBentrySTDinterwordspacing}{\spaceskip=0pt\relax}
\providecommand{\BIBentryALTinterwordstretchfactor}{4}
\providecommand{\BIBentryALTinterwordspacing}{\spaceskip=\fontdimen2\font plus
\BIBentryALTinterwordstretchfactor\fontdimen3\font minus
  \fontdimen4\font\relax}
\providecommand{\BIBforeignlanguage}[2]{{%
\expandafter\ifx\csname l@#1\endcsname\relax
\typeout{** WARNING: IEEEtran.bst: No hyphenation pattern has been}%
\typeout{** loaded for the language `#1'. Using the pattern for}%
\typeout{** the default language instead.}%
\else
\language=\csname l@#1\endcsname
\fi
#2}}
\providecommand{\BIBdecl}{\relax}
\BIBdecl

\bibitem{lang2015recommendations}
R.~M. Lang \emph{et~al.}, ``Recommendations for cardiac chamber quantification
  by echocardiography in adults: an update from the american society of
  echocardiography and the european association of cardiovascular imaging,''
  \emph{European Heart Journal-Cardiovascular Imaging}, vol.~16, no.~3, pp.
  233--271, 2015.

\bibitem{ostvik2019real}
A.~{\O}stvik, E.~Smistad, S.~A. Aase, B.~O. Haugen, and L.~Lovstakken,
  ``Real-time standard view classification in transthoracic echocardiography
  using convolutional neural networks,'' \emph{Ultrasound in medicine \&
  biology}, vol.~45, no.~2, pp. 374--384, 2019.

\bibitem{ostvik2018automatic}
A.~{\O}stvik, E.~Smistad, T.~Espeland, E.~A.~R. Berg, and L.~Lovstakken,
  ``Automatic myocardial strain imaging in echocardiography using deep
  learning,'' in \emph{International Workshop on Deep Learning in Medical Image
  Analysis}.\hskip 1em plus 0.5em minus 0.4em\relax Springer, 2018, pp.
  309--316.

\bibitem{ouyang2020video}
D.~Ouyang \emph{et~al.}, ``Video-based ai for beat-to-beat assessment of
  cardiac function,'' \emph{Nature}, vol. 580, 04 2020.

\bibitem{akerman2023automated}
\BIBentryALTinterwordspacing
A.~P. Akerman \emph{et~al.}, ``Automated echocardiographic detection of heart
  failure with preserved ejection fraction using artificial intelligence,''
  \emph{JACC: Advances}, vol.~2, no.~6, p. 100452, 2023. [Online]. Available:
  \url{https://www.sciencedirect.com/science/article/pii/S2772963X23003125}
\BIBentrySTDinterwordspacing

\bibitem{amadou2024echoapex}
A.~A. Amadou \emph{et~al.}, ``Echoapex: A general-purpose vision foundation
  model for echocardiography,'' \emph{arXiv preprint arXiv:2410.11092}, 2024.

\bibitem{echofm}
S.~Kim \emph{et~al.}, ``Echofm: Foundation model for generalizable
  echocardiogram analysis,'' \emph{IEEE Transactions on Medical Imaging}, pp.
  1--1, 2025.

\bibitem{he2022masked}
K.~He, X.~Chen, S.~Xie, Y.~Li, P.~Doll{\'a}r, and R.~Girshick, ``Masked
  autoencoders are scalable vision learners,'' in \emph{Proceedings of the
  IEEE/CVF conference on computer vision and pattern recognition}, 2022, pp.
  16\,000--16\,009.

\bibitem{zhang2024echo}
Z.~Zhang, Q.~Wu, S.~Ding, X.~Wang, and J.~Ye, ``Echo-vision-fm: A pre-training
  and fine-tuning framework for echocardiogram video vision foundation model,''
  \emph{medRxiv}, pp. 2024--10, 2024.

\bibitem{bransby2024backmix}
K.~M. Bransby, A.~Beqiri, W.-J. Cho~Kim, J.~Oliveira, A.~Chartsias, and
  A.~Gomez, ``Backmix: Mitigating shortcut learning in echocardiography with
  minimal supervision,'' in \emph{International Conference on Medical Image
  Computing and Computer-Assisted Intervention}.\hskip 1em plus 0.5em minus
  0.4em\relax Springer, 2024, pp. 570--579.

\bibitem{d2016two}
J.~D'Hooge \emph{et~al.}, ``Two-dimensional speckle tracking echocardiography:
  standardization efforts based on synthetic ultrasound data,'' \emph{European
  Heart Journal--Cardiovascular Imaging}, vol.~17, no.~6, pp. 693--701, 2016.

\bibitem{ostvik2021myocardial}
A.~{\O}stvik \emph{et~al.}, ``Myocardial function imaging in echocardiography
  using deep learning,'' \emph{ieee transactions on medical imaging}, vol.~40,
  no.~5, pp. 1340--1351, 2021.

\bibitem{evain2022motion}
E.~Evain \emph{et~al.}, ``Motion estimation by deep learning in 2d
  echocardiography: synthetic dataset and validation,'' \emph{IEEE transactions
  on medical imaging}, vol.~41, no.~8, pp. 1911--1924, 2022.

\bibitem{abulkalam2024echotracker}
M.~Abulkalam-Azad \emph{et~al.}, ``Echotracker: Advancing myocardial point
  tracking in echocardiography,'' \emph{arXiv e-prints}, pp. arXiv--2405, 2024.

\bibitem{chernyshov2024automated}
A.~Chernyshov \emph{et~al.}, ``Automated segmentation and quantification of the
  right ventricle in 2-d echocardiography,'' \emph{Ultrasound in Medicine \&
  Biology}, vol.~50, no.~4, pp. 540--548, 2024.

\bibitem{sanchez2018machine}
S.~Sanchez-Martinez \emph{et~al.}, ``Machine learning analysis of left
  ventricular function to characterize heart failure with preserved ejection
  fraction,'' \emph{Circulation: cardiovascular imaging}, vol.~11, no.~4, p.
  e007138, 2018.

\bibitem{upton2022automated}
R.~Upton \emph{et~al.}, ``Automated echocardiographic detection of severe
  coronary artery disease using artificial intelligence,'' \emph{Cardiovascular
  Imaging}, vol.~15, no.~5, pp. 715--727, 2022.

\bibitem{chiou2021ai}
Y.-A. Chiou, C.-L. Hung, and S.-F. Lin, ``Ai-assisted echocardiographic
  prescreening of heart failure with preserved ejection fraction on the basis
  of intrabeat dynamics,'' \emph{Cardiovascular Imaging}, vol.~14, no.~11, pp.
  2091--2104, 2021.

\bibitem{hathaway2022ultrasonic}
Q.~A. Hathaway, N.~Yanamala, N.~K. Siva, D.~A. Adjeroh, J.~M. Hollander, and
  P.~P. Sengupta, ``Ultrasonic texture features for assessing cardiac
  remodeling and dysfunction,'' \emph{Journal of the American College of
  Cardiology}, vol.~80, no.~23, pp. 2187--2201, 2022.

\bibitem{dorbala2021asnc}
S.~Dorbala \emph{et~al.}, ``Asnc/aha/ase/eanm/hfsa/isa/scmr/snmmi expert
  consensus recommendations for multimodality imaging in cardiac amyloidosis:
  part 1 of 2—evidence base and standardized methods of imaging,''
  \emph{Circulation: Cardiovascular Imaging}, vol.~14, no.~7, p. e000029, 2021.

\bibitem{feichtenhofer2022masked}
C.~Feichtenhofer, Y.~Li, K.~He \emph{et~al.}, ``Masked autoencoders as
  spatiotemporal learners,'' \emph{Advances in neural information processing
  systems}, vol.~35, pp. 35\,946--35\,958, 2022.

\bibitem{tong2022videomae}
Z.~Tong, Y.~Song, J.~Wang, and L.~Wang, ``Videomae: Masked autoencoders are
  data-efficient learners for self-supervised video pre-training,''
  \emph{Advances in neural information processing systems}, vol.~35, pp.
  10\,078--10\,093, 2022.

\bibitem{karaev2024cotracker}
N.~Karaev, I.~Rocco, B.~Graham, N.~Neverova, A.~Vedaldi, and C.~Rupprecht,
  ``Cotracker: It is better to track together,'' in \emph{European Conference
  on Computer Vision}.\hskip 1em plus 0.5em minus 0.4em\relax Springer, 2024,
  pp. 18--35.

\bibitem{thorley2025anatomically}
A.~Thorley \emph{et~al.}, ``Anatomically constrained transformers for cardiac
  amyloidosis classification,'' in \emph{International Workshop on Advances in
  Simplifying Medical Ultrasound}.\hskip 1em plus 0.5em minus 0.4em\relax
  Springer, 2025, pp. 248--257.

\bibitem{reynaud2021ultrasound}
H.~Reynaud, A.~Vlontzos, B.~Hou, A.~Beqiri, P.~Leeson, and B.~Kainz,
  ``Ultrasound video transformers for cardiac ejection fraction estimation,''
  in \emph{Medical Image Computing and Computer Assisted Intervention--MICCAI
  2021: 24th International Conference, Strasbourg, France, September
  27--October 1, 2021, Proceedings, Part VI 24}.\hskip 1em plus 0.5em minus
  0.4em\relax Springer, 2021, pp. 495--505.

\bibitem{mokhtari2023gemtrans}
M.~Mokhtari, N.~Ahmadi, T.~S. Tsang, P.~Abolmaesumi, and R.~Liao, ``Gemtrans: A
  general, echocardiography-based, multi-level transformer framework for
  cardiovascular diagnosis,'' in \emph{International Workshop on Machine
  Learning in Medical Imaging}.\hskip 1em plus 0.5em minus 0.4em\relax
  Springer, 2023, pp. 1--10.

\bibitem{arnab2021vivit}
A.~Arnab, M.~Dehghani, G.~Heigold, C.~Sun, M.~Lu{\v{c}}i{\'c}, and C.~Schmid,
  ``Vivit: A video vision transformer,'' in \emph{Proceedings of the IEEE/CVF
  international conference on computer vision}, 2021, pp. 6836--6846.

\bibitem{szijarto2024masked}
{\'A}.~Szij{\'a}rt{\'o} \emph{et~al.}, ``Masked autoencoders for medical
  ultrasound videos using roi-aware masking,'' in \emph{International Workshop
  on Advances in Simplifying Medical Ultrasound}.\hskip 1em plus 0.5em minus
  0.4em\relax Springer, 2024, pp. 167--176.

\bibitem{yang2025echocardmae}
X.~Yang \emph{et~al.}, ``Echocardmae: Video masked auto-encoders customized for
  echocardiography,'' in \emph{International Conference on Medical Image
  Computing and Computer-Assisted Intervention}.\hskip 1em plus 0.5em minus
  0.4em\relax Springer, 2025, pp. 171--180.

\bibitem{jamthikar2025ultrasonic}
A.~D. Jamthikar \emph{et~al.}, ``Ultrasonic texture analysis for predicting
  acute myocardial infarction,'' \emph{JACC: Cardiovascular Imaging}, 2025.

\bibitem{kirillov2023segment}
A.~Kirillov \emph{et~al.}, ``Segment anything,'' in \emph{Proceedings of the
  IEEE/CVF international conference on computer vision}, 2023, pp. 4015--4026.

\bibitem{ravishankar2023sonosam}
H.~Ravishankar, R.~Patil, V.~Melapudi, and P.~Annangi, ``Sonosam-segment
  anything on ultrasound images,'' in \emph{International Workshop on Advances
  in Simplifying Medical Ultrasound}.\hskip 1em plus 0.5em minus 0.4em\relax
  Springer, 2023, pp. 23--33.

\bibitem{ravishankar2023sonosamtrack}
H.~Ravishankar \emph{et~al.}, ``Sonosamtrack--segment and track anything on
  ultrasound images,'' \emph{arXiv preprint arXiv:2310.16872}, 2023.

\bibitem{deng2024memsam}
X.~Deng, H.~Wu, R.~Zeng, and J.~Qin, ``Memsam: taming segment anything model
  for echocardiography video segmentation,'' in \emph{Proceedings of the
  IEEE/CVF Conference on Computer Vision and Pattern Recognition}, 2024, pp.
  9622--9631.

\bibitem{porumb2021site}
M.~Porumb \emph{et~al.}, ``Site-specific automated contouring model
  generalisibiliy enhancement,'' \emph{European Heart Journal-Cardiovascular
  Imaging}, vol.~22, no. Supplement\_1, pp. jeaa356--430, 2021.

\bibitem{akbari2024beas}
S.~Akbari, M.~Tabassian, J.~Pedrosa, S.~Queir{\'o}s, K.~Papangelopoulou, and
  J.~D’Hooge, ``Beas-net: a shape-prior-based deep convolutional neural
  network for robust left ventricular segmentation in 2d echocardiography,''
  \emph{IEEE Transactions on Ultrasonics, Ferroelectrics, and Frequency
  Control}, 2024.

\bibitem{wei2020temporal}
H.~Wei \emph{et~al.}, ``Temporal-consistent segmentation of echocardiography
  with co-learning from appearance and shape,'' in \emph{Medical Image
  Computing and Computer Assisted Intervention--MICCAI 2020: 23rd International
  Conference, Lima, Peru, October 4--8, 2020, Proceedings, Part II 23}.\hskip
  1em plus 0.5em minus 0.4em\relax Springer, 2020, pp. 623--632.

\bibitem{chernyshov2025low}
A.~Chernyshov \emph{et~al.}, ``Low complexity point tracking of the myocardium
  in 2d echocardiography,'' \emph{arXiv preprint arXiv:2503.10431}, 2025.

\bibitem{vaswani2017attention}
A.~Vaswani \emph{et~al.}, ``Attention is all you need,'' \emph{Advances in
  neural information processing systems}, vol.~30, 2017.

\bibitem{harley2022particle}
A.~W. Harley, Z.~Fang, and K.~Fragkiadaki, ``Particle video revisited: Tracking
  through occlusions using point trajectories,'' in \emph{European Conference
  on Computer Vision}.\hskip 1em plus 0.5em minus 0.4em\relax Springer, 2022,
  pp. 59--75.

\bibitem{teed2020raft}
Z.~Teed and J.~Deng, ``Raft: Recurrent all-pairs field transforms for optical
  flow,'' in \emph{Computer Vision--ECCV 2020: 16th European Conference,
  Glasgow, UK, August 23--28, 2020, Proceedings, Part II 16}.\hskip 1em plus
  0.5em minus 0.4em\relax Springer, 2020, pp. 402--419.

\bibitem{loshchilov2016sgdr}
I.~Loshchilov and F.~Hutter, ``Sgdr: Stochastic gradient descent with warm
  restarts,'' \emph{arXiv preprint arXiv:1608.03983}, 2016.

\bibitem{goyal2017accurate}
P.~Goyal, ``Accurate, large minibatch sg d: training imagenet in 1 hour,''
  \emph{arXiv preprint arXiv:1706.02677}, 2017.

\bibitem{Sharmila}
S.~Dorbala \emph{et~al.}, ``Asnc/aha/ase/eanm/hfsa/isa/scmr/snmmi expert
  consensus recommendations for multimodality imaging in cardiac amyloidosis:
  Part 2 of 2—diagnostic criteria and appropriate utilization,''
  \emph{Circulation: Cardiovascular Imaging}, vol.~14, no.~7, p. e000030, 2021.

\bibitem{asch2022human}
F.~M. Asch \emph{et~al.}, ``Human versus artificial intelligence--based
  echocardiographic analysis as a predictor of outcomes: an analysis from the
  world alliance societies of echocardiography covid study,'' \emph{Journal of
  the American Society of Echocardiography}, vol.~35, no.~12, pp. 1226--1237,
  2022.

\bibitem{alessandrini2017realistic}
M.~Alessandrini \emph{et~al.}, ``Realistic vendor-specific synthetic ultrasound
  data for quality assurance of 2-d speckle tracking echocardiography:
  Simulation pipeline and open access database,'' \emph{IEEE transactions on
  ultrasonics, ferroelectrics, and frequency control}, vol.~65, no.~3, pp.
  411--422, 2017.

\end{thebibliography}
\end{document}